\documentclass[conference]{IEEEtran}
\IEEEoverridecommandlockouts
\usepackage{cite}
\usepackage{amsmath,amssymb,amsfonts}
\usepackage{algorithmic}
\usepackage{textcomp}
\usepackage{silence}
\usepackage{colortbl}
\usepackage{hhline} 
\usepackage{subcaption} 
\usepackage{multirow}
\usepackage{float} 
\usepackage{hyperref}
\usepackage{fontawesome5}
\usepackage{amssymb}
\usepackage{pifont}

\newcommand{\hcmark}{\textcolor{green!50!black}{\faCheck}}
\newcommand{\hxmark}{\textcolor{red}{\faTimes}}

\usepackage[table]{xcolor} 
\definecolor{lightgray}{gray}{0.9}
\definecolor{darkgray}{gray}{0.8}
\definecolor{darkgreen}{RGB}{0,100,0}
\definecolor{mygreen1}{RGB}{240, 245, 230}
\definecolor{mygblue1}{RGB}{217, 243, 253}
\definecolor{mygorange1}{RGB}{254, 239, 227}
\definecolor{mygreen2}{RGB}{92, 187, 86}
\definecolor{mygblue2}{RGB}{2, 183, 232}
\definecolor{mygorange2}{RGB}{247, 148, 29}
\usepackage{fancyhdr}
\fancypagestyle{firstpage}{
    \fancyhf{} 
    \fancyfoot[L]{\footnotesize \textbf{Disclaimer:} This is the accepted version of a paper published in IEEE ICNS-2025. The DOI will be updated once the final version is available.  

© 2024 IEEE. Personal use of this material is permitted. Permission from IEEE must be obtained for all other uses, including reprinting/republishing, creating new collective works, and reuse of any copyrighted component of this work.}
}

\usepackage{graphicx} 

\def\BibTeX{{\rm B\kern-.05em{\sc i\kern-.025em b}\kern-.08em
    T\kern-.1667em\lower.7ex\hbox{E}\kern-.125emX}}
    
\begin{document}

\title{Synthetic Flight Data Generation Using Generative Models\\
\thanks{Funded by SESAR 3 Joint Undertaking, co-funded by the European Union.}
}
\author{\IEEEauthorblockN{Karim Aly}
\IEEEauthorblockA{\textit{Faculty of Aerospace Engineering} \\
\textit{Delft University of Technology (TU Delft)}\\
Delft, The Netherlands\\
k.y.s.b.aly@tudelft.nl}
\and
\IEEEauthorblockN{Alexei Sharpanskykh}
\IEEEauthorblockA{\textit{Faculty of Aerospace Engineering} \\
\textit{Delft University of Technology (TU Delft)}\\
Delft, The Netherlands\\
o.a.sharpanskykh@tudelft.nl}
}

\maketitle
\thispagestyle{firstpage} 
\begin{abstract}
The increasing adoption of synthetic data in aviation research offers a promising solution to data scarcity and confidentiality challenges. This study investigates the potential of generative models to produce realistic synthetic flight data and evaluates their quality through a comprehensive four-stage assessment framework. The need for synthetic flight data arises from their potential to serve as an alternative to confidential real-world records and to augment rare events in historical datasets. These enhanced datasets can then be used to train machine learning models that predict critical events, such as flight delays, cancellations, diversions, and turnaround times. Two generative models, Tabular Variational Autoencoder (TVAE) and Gaussian Copula (GC), are adapted to generate synthetic flight information and compared based on their ability to preserve statistical similarity, fidelity, diversity, and predictive utility. Results indicate that while GC achieves higher statistical similarity and fidelity, its computational cost hinders its applicability to large datasets. In contrast, TVAE efficiently handles large datasets and enables scalable synthetic data generation. The findings demonstrate that synthetic data can support flight delay prediction models with accuracy comparable to those trained on real data. These results pave the way for leveraging synthetic flight data to enhance predictive modeling in air transportation. 

\end{abstract}


\section{Introduction}
\label{sec:introduction}
The aviation industry increasingly relies on artificial intelligence (AI) and machine learning (ML) to optimize air transport operations, enhance efficiency, reduce delays, and improve decision-making processes. These technologies enable airlines, airports, and air traffic controllers to make data-driven decisions that improve scheduling, fuel efficiency, and passenger experiences. However, a key challenge remains: the scarcity of comprehensive, high-quality datasets due to limited data collection possibilities, strict data privacy regulations, commercial competition, proprietary restrictions, and regulatory barriers. Many critical flight-related events, such as delays, diversions, and cancellations, occur infrequently, making them rare in historical datasets. This scarcity complicates the study and prediction of such events, a challenge known as class imbalance in machine learning. These constraints hinder the development of accurate predictive models and the generalization of machine learning-based solutions, ultimately limiting their applicability to real-world aviation scenarios. A promising approach to addressing these challenges is synthetic data generation (SDG), which creates artificial datasets that closely replicate real-world scenarios while maintaining essential statistical properties \cite{figueira2022survey}. SDG can not only supplement existing datasets but also help augment machine learning training data, addressing the issues of class imbalance and the underrepresentation of rare events.

Recent advancements in generative AI have produced powerful models capable of generating high-fidelity synthetic tabular data. Notable approaches include probabilistic models like the Gaussian Copula (GC) \cite{sdv} and deep learning-based methods such as the Tabular Variational Autoencoder (TVAE) \cite{xu2019modeling}. These techniques use different strategies to model real-world data distributions. GC relies on statistical modeling, while TVAE employs neural network architecture. Despite their promise, the application of these approaches in the area of air transportation remains largely underexplored, particularly regarding their ability to preserve data fidelity and enhance predictive modeling. Although synthetic data offers clear benefits, its effectiveness in aviation-related tasks remains uncertain, highlighting the need for rigorous benchmarking against real data through statistical and machine learning assessments to validate its applicability in this domain. Furthermore, aviation data poses unique challenges, such as complex temporal dependencies and operational constraints, which necessitate careful model selection and evaluation. 

Since access to key attributes of European flight data, such as flight schedules, statuses, delays, and diversion information, is restricted, this study aims to evaluate the feasibility of using generative models to produce realistic synthetic flight information. Specifically, we examine the potential of TVAE and GC in generating synthetic flight data, with a particular focus on their impact on flight delay prediction accuracy. To achieve this, we conduct five experiments using different sets of features as input to the generative models, aiming to identify the optimal set of features and data types that allow the model to learn the underlying characteristics of real-world flight data.

We propose a four-stage evaluation framework that assesses the statistical similarity, fidelity, diversity, and predictive performance of the generated data. Our findings reveal a trade-off between the size of the synthetic data that can be produced and the utility of these data for predictive tasks. While GC demonstrates superior statistical similarity, its high computational demand limits scalability, resulting in smaller synthetic datasets that may not be ideal for training predictive models due to the underrepresentation of various flight patterns. In contrast, TVAE offers greater scalability, capable of being trained on large datasets that include all flight patterns and generating large synthetic datasets that retain these patterns. However, it exhibits higher sensitivity to feature selection and data types.

Despite these challenges, our results suggest that synthetic data can effectively support predictive modeling in aviation, providing a viable solution to the data access challenges that hinder research in air transportation. By enabling the development of more robust machine learning models, synthetic data generation can help address key limitations associated with real-world datasets.

The remainder of this paper is structured as follows: Section~\ref{sec:related_work} reviews key methodologies for synthetic tabular data generation. Section~\ref{sec:methodology} describes the methodological framework employed to generate synthetic flight data using TVAE and GC, covering all stages from preprocessing the raw historical data to evaluating the synthetically generated datasets. Section~\ref{sec:results} presents the results of our comparative experiments, assessing both the statistical fidelity and the predictive utility of the generated data across multiple evaluation metrics. Section~\ref{sec:discussion} highlights key insights, discusses limitations, and explores practical considerations for deploying synthetic data in air transport applications. Finally, Section~\ref{sec:conclusions_and_future_work} provides concluding remarks and outlines future research directions, focusing on further improving the quality and realism of synthetic flight data.

\section{Related work}
\label{sec:related_work}
Synthetic data generation for tabular datasets has advanced considerably, evolving from traditional statistical techniques to sophisticated deep learning-based models. This section reviews key methodologies, highlighting their applications and limitations while identifying gaps in the literature concerning the utilization of synthetic flight data in air transportation and air traffic management (ATM).
 
Statistical and probabilistic methods like Gaussian Copulas (GC) remain powerful tools for synthetic data generation, effectively decomposing multivariate distributions into marginal distributions and dependency structures to capture complex relationships efficiently \cite{sdv}. Their flexibility has been further enhanced through extensions such as Archimedean and Vine Copulas, which allow for more adaptable hierarchical dependency modeling in high-dimensional datasets. Additionally, oversampling techniques like the Synthetic Minority Over-sampling Technique (SMOTE) \cite{chawla2002smote} and the Adaptive Synthetic (ADASYN) sampling approach \cite{he2008adasyn} were originally designed to address class imbalances by generating synthetic samples for minority classes, but have since been repurposed for general synthetic data generation.

Deep learning has revolutionized synthetic data generation, particularly through Variational Autoencoders (VAEs) \cite{Kingma2013}, which have been adapted for tabular data via architectures like the Tabular Variational Autoencoder (TVAE) \cite{xu2019modeling}. TVAEs introduce modifications that accommodate mixed categorical and continuous variables, improving their ability to preserve complex feature relationships. Similarly, Generative Adversarial Networks (GANs) have gained traction since their introduction by Goodfellow et al. \cite{goodfellow2014generative}, leading to tabular adaptations such as Table-GAN and TGAN \cite{park2018data, xu2018synthesizing}. Notably, CTGAN \cite{xu2019modeling} effectively addresses challenges associated with categorical variables through conditional generation and mode-specific normalization, while medGAN \cite{choi2017generating} pioneered the use of GANs for discrete electronic health records. Other refinements, such as VeeGAN \cite{srivastava2017veegan}, introduce mechanisms to mitigate mode collapse, further improving the robustness of synthetic data generation.

Although synthetic data is widely used in domains such as finance, healthcare, cybersecurity, computer vision, and manufacturing, its adoption in air transportation remains limited. Most existing research focuses on downstream machine learning tasks, such as predicting flight delays, cancellations, diversions, and turnaround times, relying solely on historical data. However, these datasets often suffer from significant class imbalances, as critical events like cancellations and diversions occur far less frequently than on-time flights, making accurate predictions challenging. Rather than augmenting these datasets with synthetic flight records, research efforts have primarily concentrated on refining predictive models while overlooking the fundamental limitations of the data itself. Beyond augmenting rare events in historical datasets, the potential for conditionally generating synthetic flight data to simulate hypothetical scenarios—such as extreme weather conditions or congested airspace—remains largely unexplored.

Prior work that references synthetic flight data has primarily focused on generating synthetic flight trajectories \cite{Liu2021DeepLearningAidedPR, Zhang2024AnEA, Wijnands2024}. In contrast, this study defines flight data more broadly, encompassing key flight attributes such as flight number, airline and aircraft information, origin and destination airports, scheduled and actual departure and arrival times, air time, and operational flight logs indicating delays, cancellations, and diversions. Many of these features, particularly for European flights, are restricted or unavailable in public datasets, limiting researchers' ability to develop robust machine learning models. This study seeks to bridge this gap by demonstrating how generative models can supplement real-world flight data, providing a scalable and practical solution to data scarcity challenges in ATM.

Despite the growing number of generative models and their continuous evolution, no single approach can be universally regarded as the best. The performance of synthetic data generation methods is highly dataset-dependent, and in many cases, well-established techniques still outperform newer variants on specific datasets. Historical flight data encompasses flights between numerous airports at different times of the day and under varying conditions, requiring generative models capable of capturing this complexity. These models must generate synthetic data that reflects the full variability of real-world flight patterns without overfitting to a specific subset. For this reason, we adopt Gaussian Copulas (GC) from statistical modeling and the Tabular Variational Autoencoder (TVAE) from deep learning to generate synthetic flight data. While each method brings different advantages to the analysis, both are recognized for their stability and reduced susceptibility to mode collapse, a common issue in GAN-based approaches, making them well-suited for our application.

\section{Methodology}
\label{sec:methodology}
\begin{figure}[H]
    \centering
    \includegraphics[width=0.48\textwidth]{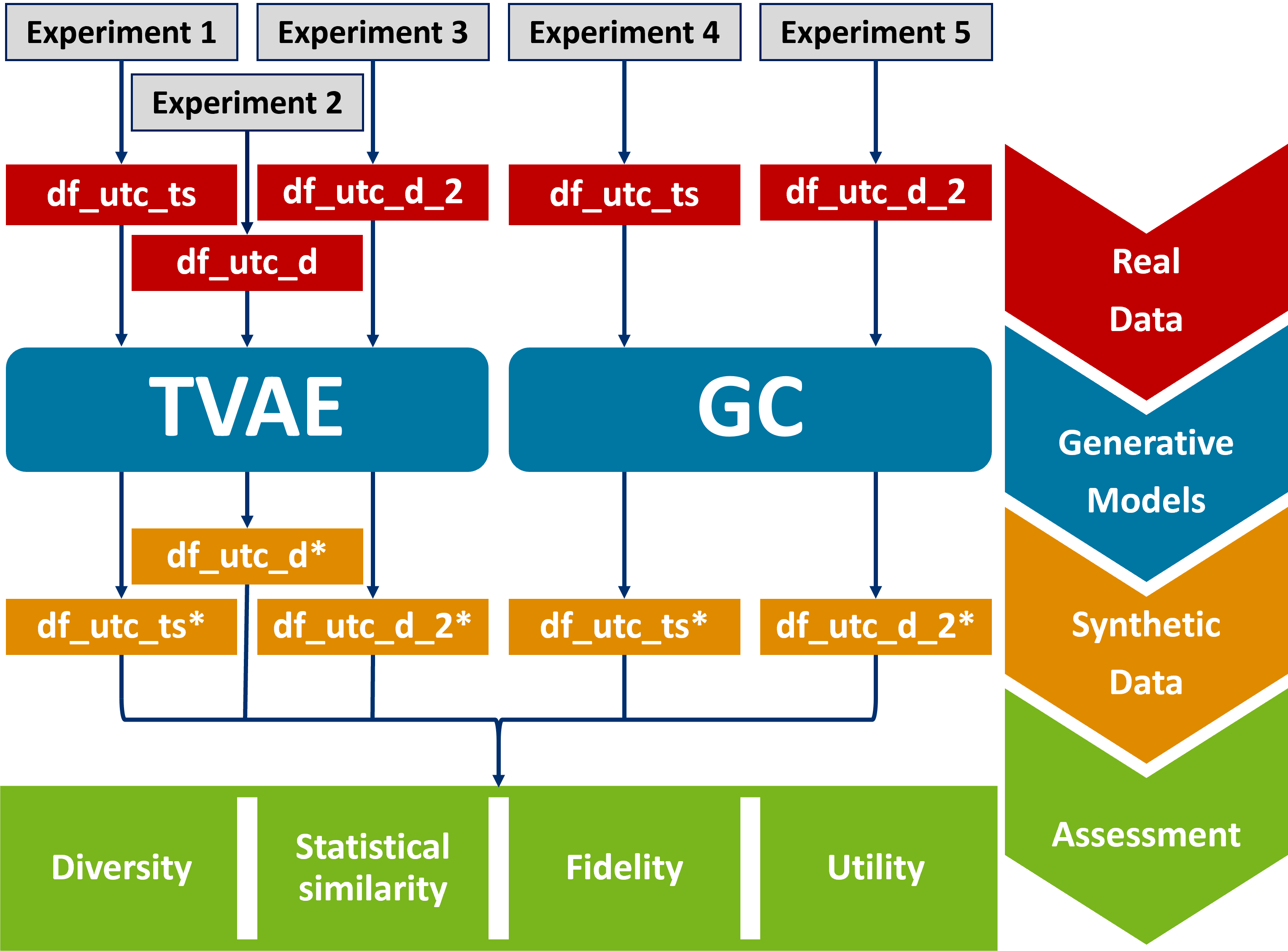} 
    \caption{Overview of the analysis framework.}
    \label{fig:methodology}
\end{figure}

This section describes our analysis framework, covering data collection, preprocessing, and feature engineering. It also details the generative models used in five experiments with three different sets of input features. Finally, we outline the four key evaluation criteria applied to assess the experimental results. Fig.~\ref{fig:methodology} provides an overview of the entire analysis process.

\subsection{Data and Preprocessing}
\label{sec:data}
This study uses the publicly available ``TranStats Database for Airline On-Time Performance" \cite{TranStats} from the Bureau of Transportation Statistics (BTS) \cite{bts}. The data covers U.S. domestic flights and provides detailed information on flight delays, cancellations, diversions, and their causes. This level of detail makes it a valuable resource for modeling various use cases in air transportation. 

To make the task more challenging for the generative models, we used flight data for all arrivals and departures in New York State during January 2023, rather than limiting the data to flights between two specific airports. This resulted in a dataset with 109 features (columns) and approximately 61,000 flights (rows), spanning 113 airports and 508 routes. 

The data underwent extensive exploratory analysis, preprocessing, and feature engineering to ensure clean and well-structured input for the generative models. Randomly missing values, such as missing ``Tail number", were removed from the dataset. However, other missing values, like arrival times for cancelled flights, were retained, as they carry meaningful information and were incorporated into the training of the generative model.

Departure and arrival time features were initially represented as local times in integer format (HHMM) without the associated date components. During the preprocessing phase, the date component from the ``Flight Date" feature was combined with the ``Scheduled Departure Time". Using additional duration-based features, such as ``Air Time (min)" and ``Scheduled Elapsed Time (min)", the appropriate date components were calculated and appended to the remaining time features. To ensure consistency, all time features were converted into timezone-aware datetime objects based on the time zones of the origin and destination airports. Finally, they were standardized to Coordinated Universal Time (UTC) to achieve a unified temporal representation throughout the dataset.

Based on their relevance to this study, the number of features was reduced to 30, encompassing categorical, numerical, and datetime features, along with relational attributes that can be derived from others. This combination of different data types and the complex relationships among them presents challenges for synthetic data generation. Directly including all 30 features as input to the generative models may disrupt inherent dependencies, leading to inconsistencies. For instance, the ``Origin Airport ID" could be incorrectly paired with an unrelated ``Origin City", or the ``Scheduled Elapsed Time (min)" might not align with the difference between ``Scheduled Arrival Time UTC" and ``Scheduled Departure Time UTC".  

To preserve these dependencies and enhance the fidelity of the generated flight information, the dataset was organized into three distinct DataFrames: \textit{``df\_utc\_ts"}, \textit{``df\_utc\_d"}, and \textit{``df\_utc\_d\_2"}. Each DataFrame incorporates different sets of features as input to the generative models. These DataFrames were used in five generation tests, as described in Section~\ref{sec:experiments}, to assess the impact of various feature combinations on the quality of the generated data.

Table~\ref{tab:features} provides an overview of the content of each DataFrame, specifying the features included as input to the generative models and the relational features that can be computed or inferred after generation. The first DataFrame primarily consists of time-related features represented as timestamps (datetime format), while the second and third focus mainly on time durations in minutes (numeric format), with a limited number of timestamps included. This structured approach allows for a systematic analysis of how different feature representations influence the ability of the generative models to learn and replicate the underlying patterns in real-world flight data.

By testing multiple feature configurations, we aim to determine the optimal set of attributes that maximize the realism and fidelity of the generated data while preserving essential relationships among features. Ensuring these dependencies remain intact is crucial for maintaining the operational correctness of synthetic flight records, particularly for downstream machine learning tasks such as flight delay prediction. The last column of Table~\ref{tab:features} specifies the input features used in the predictive models discussed in Section~\ref{sec:evaluation}, which serve as a benchmark for assessing the practical utility of the synthetic data in real-world aviation scenarios.

\renewcommand{\arraystretch}{1.2} 
\setlength{\tabcolsep}{8pt} 
\begin{table}[H]
\caption{\centering Features used in this analysis, categorized as: included (\hcmark), excluded (\hxmark), or calculated post-generation (\faCalculator).}

\begin{center}
\resizebox{0.47\textwidth}{!}{%
\begin{tabular}{||l|c|c|c|c||}
\hline \rowcolor{darkgray}
\textbf{Features} & \textbf{df\_utc\_ts} & \textbf{df\_utc\_d} & \textbf{df\_utc\_d\_2} & \textbf{df\_prediction} \\ \hline \hline
Unique Carrier Code & \hcmark & \hcmark & \hcmark & \hcmark \\ \hline
Tail Number & \hcmark & \hcmark & \hcmark & \hcmark \\ \hline
Origin Airport ID & \hcmark & \hcmark & \hcmark & \hxmark \\ \hline
ICAO Origin Airport & \faCalculator & \faCalculator & \faCalculator & \hcmark \\ \hline
Origin City & \faCalculator & \faCalculator & \faCalculator & \hxmark \\ \hline
Origin State Code & \faCalculator & \faCalculator & \faCalculator & \hxmark \\ \hline
Origin State Name & \faCalculator & \faCalculator & \faCalculator & \hxmark \\ \hline
Destination Airport ID & \hcmark & \hcmark & \hcmark & \hxmark \\ \hline
ICAO Destination Airport & \faCalculator & \faCalculator & \faCalculator & \hcmark \\ \hline
Destination City & \faCalculator & \faCalculator & \faCalculator & \hxmark \\ \hline
Destination State Code & \faCalculator & \faCalculator & \faCalculator & \hxmark \\ \hline
Destination State Name & \faCalculator & \faCalculator & \faCalculator & \hxmark \\ \hline
Quarter & \faCalculator & \faCalculator & \faCalculator & \hcmark \\ \hline
Day of Week & \faCalculator & \faCalculator & \faCalculator & \hcmark \\ \hline
Scheduled Departure Time UTC & \hcmark & \hcmark & \hcmark & \hcmark \\ \hline
Actual Departure Time UTC & \hcmark & \faCalculator & \hcmark & \hcmark \\ \hline
Departure $\Delta$T (min) & \faCalculator & \hcmark & \hcmark & \hcmark \\ \hline
Departure Delay Label & \faCalculator & \faCalculator & \faCalculator & \hxmark \\ \hline
Taxi Out Time (min) & \faCalculator & \hcmark & \hcmark & \hcmark \\ \hline
Wheels Off Time UTC & \hcmark & \faCalculator & \faCalculator & \hcmark \\ \hline
Wheels On Time UTC & \hcmark & \faCalculator & \faCalculator & \hxmark \\ \hline
Taxi In Time (min) & \faCalculator & \faCalculator & \hcmark & \hxmark \\ \hline
Scheduled Arrival Time UTC & \hcmark & \faCalculator & \faCalculator & \hcmark \\ \hline
Actual Arrival Time UTC & \hcmark & \faCalculator & \faCalculator & \hxmark \\ \hline
Arrival $\Delta$T (min) & \faCalculator & \faCalculator & \hcmark & \textcolor{green!50!black}{\textbf{Target}} \\ \hline
Arrival Delay Label & \faCalculator & \faCalculator & \faCalculator & \hxmark \\ \hline
Scheduled Elapsed Time (min) & \faCalculator & \hcmark & \hcmark & \hcmark \\ \hline
Actual Elapsed Time (min) & \faCalculator & \hcmark & \hcmark & \hxmark \\ \hline
Air Time (min) & \faCalculator & \hcmark & \hcmark & \hxmark \\ \hline
Distance (miles) & \faCalculator & \faCalculator & \faCalculator & \hcmark \\ \hline
\end{tabular}
}
\label{tab:features}
\end{center}
\end{table}

\subsection{Generative Models}
\label{sec:gen_models}
In this study, we employed the Tabular Variational Autoencoder (TVAE) and Gaussian Copula (GC) models to generate synthetic flight data, including trip logs that indicate whether flights departed or arrived on time or were delayed.

One of the primary challenges in generating tabular data is managing the variety of feature types (e.g., numerical, categorical, datetime) and handling missing values that might carry additional information about the dataset. Since synthetic data must replicate the structure of the original data, any missing values in the original dataset must be mirrored in the generated data. Both TVAE and GC models assume that the data columns are fully populated with numerical values. In cases where these assumptions do not hold, a preprocessing step is required. This step modifies the data by transforming columns of one type into one or more columns of another type, as outlined in Table~\ref{tab:conversions}.

To address columns with missing values, each of such columns is split into two: a column of the same type, where missing values are filled by randomly selecting non-missing values from the same column, and a categorical column indicating whether the original data was present (``Yes") or missing (``No") for each row. This approach ensures that the original column is fully populated, while also accounting for the presence of missing values in the original dataset \cite{sdv}.

\renewcommand{\arraystretch}{1.2} 
\setlength{\tabcolsep}{8pt} 
\begin{table}[H]
\caption{\centering Data conversion during preprocessing to handle non-numeric and missing values, adapted from \cite{sdv}.}
\begin{center}
\resizebox{0.47\textwidth}{!}{%
\begin{tabular}{||l|l||}
\hline \rowcolor{darkgray} \textbf{Original Column Type} & \textbf{Replaced Column(s) Type} \\
\hline  \hline Categorical & Number \\
\hline Datetime & Number \\
\hline Number w/Missing Values & Number \& Categorical \\
\hline Categorical w/Missing Values & Categorical \& Categorical \\
\hline Datetime w/Missing Values & Datetime \& Categorical \\
\hline
\end{tabular}
}
\label{tab:conversions}
\end{center}
\end{table}

The TVAE is a deep learning model designed to extend the functionality of traditional autoencoders by incorporating probabilistic modeling tailored to tabular data \cite{Yang2023TowardsAC}. The model includes an encoder neural network that maps the input data, $x$, into a probabilistic distribution over the latent space, $z$, represented as $q(z | x)$. A decoder network reconstructs the data as the conditional distribution $p(x | z)$. This process allows the TVAE to learn the underlying patterns and relationships within the data, enabling the generation of synthetic data that closely mimics the original data \cite{Shen2024TowardsAF}. The objective function of TVAE is to maximize the Evidence Lower Bound (ELBO) on the log-likelihood of data  \cite{al2018intriguing}, denoted by:

\begin{equation}
\label{eq:elbo}
ELBO = \mathbb{E}_{z \sim q(z|x)}[\log p(x|z)] - D_{\text{KL}}(q(z|x) \| p(z))
\end{equation}

Maximizing the data reconstruction likelihood $\log p(x|z)$, represented as the expectation $\mathbb{E}_{z \sim q(z|x)}[\log p(x|z)]$, ensures that the decoder accurately reconstructs the input from the latent representation. Simultaneously, minimizing the Kullback-Leibler (KL) divergence $D_{\text{KL}}(q(z|x) \| p(z))$ ensures that the approximate posterior $q(z|x)$ aligns closely with the prior $p(z)$, promoting a coherent and structured latent space \cite{Kingma2013, Akrami2020RobustVA}.

Through our analysis, we adapted the same TVAE model structure as in \cite{xu2019modeling}. The TVAE was trained for 300 epochs using Adam optimizer with a learning rate 1e-3 and ELBO loss \eqref{eq:elbo}.

The Gaussian Copula is a statistical model that allows for the modeling of complex dependencies between variables while preserving their marginal distributions \cite{nelsen2006introduction}. It operates by transforming the marginal distributions of the data into uniform distributions through their cumulative distribution functions (CDFs), and then combining them using a Gaussian Copula function \cite{Khosravi2024BinaryGC, Asghar2019DifferentiallyPR, Jiang2022MeasuringRR}. The Gaussian Copula is defined by the correlation structure of the underlying multivariate normal distribution. Specifically, for random variables $X_1, X_2, \dots, X_d$ with marginals $F_1(x_1), F_2(x_2), \dots, F_d(x_d)$, the copula $C_\theta$ captures the joint distribution as:

\begin{multline}
C_\theta\left(F_1(x_1), F_2(x_2), \dots, F_d(x_d)\right) = \\
\Phi_\theta\left( \Phi^{-1}(F_1(x_1)), \Phi^{-1}(F_2(x_2)), \dots, \Phi^{-1}(F_d(x_d)) \right)
\end{multline}

where $\Phi_\theta$ represents the joint CDF of the multivariate normal distribution with correlation matrix $\theta$, and $\Phi^{-1}$ is the inverse of the standard normal CDF. This copula model enables the generation of synthetic data that captures realistic dependencies while maintaining the marginal distributions of the variables.

This research combines Gaussian Copulas with Kernel Density Estimation (KDE) from \cite{sdv_copulas_index}, using a Gaussian kernel to estimate the marginal distributions of each variable. This method allowed us to model the interdependencies between variables and generate synthetic data that closely reflects the joint distribution of the original dataset.

\subsection{Experiments}
\label{sec:experiments}
Five experiments were conducted to systematically evaluate the impact of input data types and feature selection on the quality of synthetic flight information generated by Tabular Variational Autoencoder (TVAE) and Gaussian Copula (GC). The experiments were designed as follows:

\begin{itemize}
    \item \textbf{\textit{Experiment 1: TVAE with df\_utc\_ts}}
    \item \textbf{\textit{Experiment 2: TVAE with df\_utc\_d}}
    \item \textbf{\textit{Experiment 3: TVAE with df\_utc\_d\_2}}
    \item \textbf{\textit{Experiment 4: GC with df\_utc\_ts}}
    \item \textbf{\textit{Experiment 5: GC with df\_utc\_d\_2}}
\end{itemize}

Table~\ref{tab:data_size} presents the input data sizes used to train the generative models in the different experiments, along with the size of the synthetic datasets sampled from the learned distributions. TVAE was trained on approximately 61,000 flights, whereas GC was limited to 5,000 flights due to memory constraints. Similarly, sampling with GC is computationally expensive, so only 5,000 flights were generated.

\renewcommand{\arraystretch}{1.2} 
\setlength{\tabcolsep}{8pt} 
\begin{table}[H]
\caption{\centering Data sizes}
\begin{center}
\resizebox{0.47\textwidth}{!}{%
\begin{tabular}{||l||c|c|c|c||}
\hline \rowcolor{darkgray} \textbf{Experiments} & \textbf{Input (real)} & \textbf{Sampled (syn.)} & \textbf{Reconstructed} & \textbf{Cleaned} \\
\hline  
\hline \cellcolor{darkgray} Experiment 1 & (60767, 10) & (60000, 10) & (60000, 30) & (24422, 30) \\
\hline \cellcolor{darkgray} Experiment 2 & (60767, 10) & (60000, 10) & (60000, 30) & (50954, 30) \\
\hline \cellcolor{darkgray} Experiment 3 & (60767, 13) & (60000, 13) & (60000, 30) & (51685, 30) \\
\hline \cellcolor{darkgray} Experiment 4 & (5000, 10) &  (5000, 10)  & (5000, 30)  & (2433, 30) \\
\hline \cellcolor{darkgray} Experiment 5 & (5000, 13) &  (5000, 13)  & (5000, 30)  & (2414, 30) \\

\hline 
\end{tabular}
}
\label{tab:data_size}
\end{center}
\end{table}

After generation, we reconstructed relational features inferred from other variables and applied rejection sampling to remove synthetic routes that did not exist in historical data. The cleaned synthetic datasets were used in the evaluation framework described in the next section. 

\subsection{Evaluation Framework}
\label{sec:evaluation}
Evaluating synthetic data is more critical than generating it, as unreliable synthetic data can lead to incorrect conclusions. Without rigorous validation, synthetic data cannot be trusted to be used for downstream tasks. A standard evaluation step is to assess the validity and structure of the generated data by ensuring that the number of generated features matches the real data, continuous values remain within the observed min/max range, and discrete values correspond to the original categories. Beyond validity, the evaluation framework focuses on four aspects: data diversity, to ensure the synthetic dataset captures the variability of the real data; statistical similarity, by comparing distributions and correlations; fidelity, measuring how well synthetic samples preserve patterns from real data; and utility, determining how well models trained on synthetic data perform in comparison to those trained on real data.

The diversity assessment included applying Principal Component Analysis (PCA) to project both real and synthetic data into a two-dimensional space \cite{wold1987principal}, allowing for a visual evaluation of whether the synthetic data captured the different distributions and clusters present in the real data. Additionally, the class balance was examined by inspecting the arrival and departure delay labels to ensure that the synthetic data maintained a similar or nearly identical ratio of on-time to delayed flights as the real data. Maintaining diversity is crucial, as insufficient variability in synthetic data may lead to biased or unrepresentative models in downstream tasks. 

The statistical assessment of the synthetic data was performed both visually and numerically. Visually, we compared the marginal and bivariate distribution plots of the real and synthetic data to identify discrepancies in statistical patterns. To compare individual distributions numerically, we used the Total Variation Distance (TVD) to quantify the divergence between the probability distributions of boolean and categorical columns \cite{Knoblauch2020RobustBI}, and for numerical and datetime columns, we calculated the statistical similarity using the Kolmogorov-Smirnov test \cite{viehmann2021numerically}, which measures the difference between the cumulative distribution functions (CDFs) of the real and synthetic datasets. To compare relationships between feature pairs, Correlation Similarity \cite{sdmetrics_correlationsimilarity} and Contingency Similarity \cite{sdmetrics_contingencysimilarity} were employed.

To assess the fidelity of the synthetic data, seven classifiers were trained for a binary classification task to distinguish between real and synthetic data. Each model brings unique strengths to the assessment: Random Forest and Gradient Boosting capture complex, non-linear relationships \cite{breiman2001random, friedman2001greedy}; K-Nearest Neighbors (KNN) and Decision Trees are effective for detecting local patterns \cite{cover1967nearest, breiman2017classification}; Naive Bayes provides insights into probabilistic dependencies \cite{lewis1998naive}; Logistic Regression models linear relationships \cite{cox1958regression}; and Stochastic Gradient Descent (SGD) is well-suited for handling large-scale data \cite{SGDClassifier}. Stratified KFold cross-validation was employed to ensure class distribution was preserved across folds \cite{Ahmadi2024ACS}, with five splits and shuffling to enhance model robustness. Performance was evaluated by averaging accuracy and F1 scores across all classifiers \cite{sokolova2009systematic}, providing a comprehensive measure of their ability to differentiate between the real and synthetic datasets. A lower classification accuracy indicates higher similarity between synthetic and real data, as the models struggle to distinguish between them. 

The utility of the synthetic data was assessed by evaluating whether it preserved or enhanced the predictive characteristics compared to real data. Accurate prediction from models trained on synthetic data indicates that it can be reliably used as a substitute for real data in downstream tasks and decision-making. For this reason, sixteen regression models were trained to predict flight arrival delays in minutes, with each offering distinct advantages for the evaluation. Similar to the choices of the classification models, the selection of the regression models was strategically diverse, encompassing algorithms capable of capturing both linear and non-linear relationships, handling high-dimensional feature spaces, and identifying local patterns within the data. Additionally, several ensemble learning techniques were incorporated to leverage their enhanced predictive performance, while other models were specifically chosen to assess the impact of dimensionality reduction. Each experiment from Section~\ref{sec:experiments}, included two testing scenarios: (1) Train-Real-Test-Real (TRTR), which established the baseline performance using historical data, and (2) Train-Synthetic-Test-Real (TSTR), which evaluated the utility of synthetic data for model training. The input features used for these regression models are outlined in the last column of Table~\ref{tab:features}, with a careful exclusion of time-related variables that could directly infer the actual arrival time at destination airports, ensuring that the models were trained on information that did not directly relate to the target variable. Model performance was quantified using three complementary metrics: Mean Absolute Error (MAE), Root Mean Square Error (RMSE), and R-squared (R²) \cite{willmott2005advantages, mentch2016quantifying, cameron1997r}. These evaluation metrics were averaged across all models to provide a robust assessment of synthetic data quality. The comparative analysis of the TRTR and TSTR results offers valuable insights into the machine learning utility of synthetic data, as well as its potential as a replacement for historical data in downstream predictive tasks.

Due to the varying data sizes used for training TVAE (in Experiments 1, 2, and 3) and GC (in Experiments 4 and 5), along with the differences in the sizes of the sampled data (see Table~\ref{tab:data_size}), we expect the PCA plots in Section~\ref{sec:diversity_assessment} to exhibit distinct clustering patterns. Likewise, we anticipate notable differences in the distribution plots presented in Section~\ref{sec:statistical_assessment}. Moreover, these discrepancies in data size are likely to impact the overall utility of the synthetic datasets for machine learning predictive tasks, underscoring the trade-offs between scalability and utility in generative modeling.

\section{Results}
\label{sec:results}
This section evaluates the quality of the synthetic flight information generated by the adapted Tabular Variational Autoencoder (TVAE) and Gaussian Copula (GC) models through the five experiments described in Section~\ref{sec:experiments}. The assessment follows the evaluation framework detailed in Section~\ref{sec:evaluation}.

\subsection{Diversity Assessment}
\label{sec:diversity_assessment}
\begin{figure}[H]
    \centering
    \begin{minipage}{0.155\textwidth} 
        \centering
        \includegraphics[width=\textwidth]{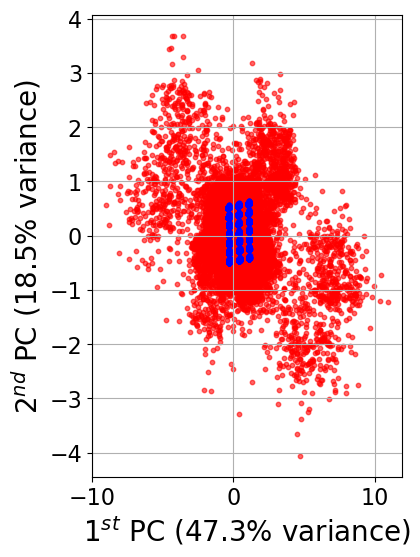} 
        \subcaption{\centering Experiment 1}
        \label{fig:pca_exp_1}
    \end{minipage}
    \hfill
    \begin{minipage}{0.155\textwidth} 
        \centering
        \includegraphics[width=\textwidth]{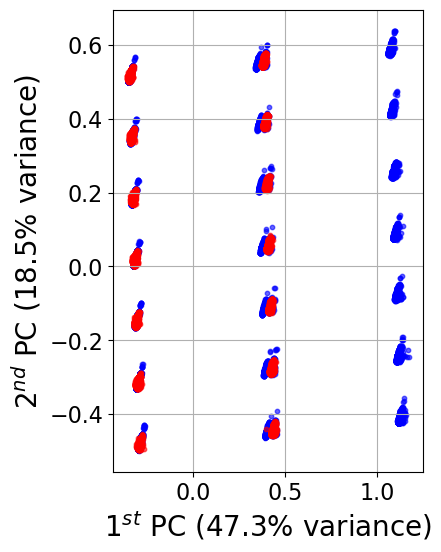} 
        \subcaption{\centering Experiment 2}
        \label{fig:pca_exp_2}
    \end{minipage}
    \hfill
    \begin{minipage}{0.155\textwidth} 
        \centering
        \includegraphics[width=\textwidth]{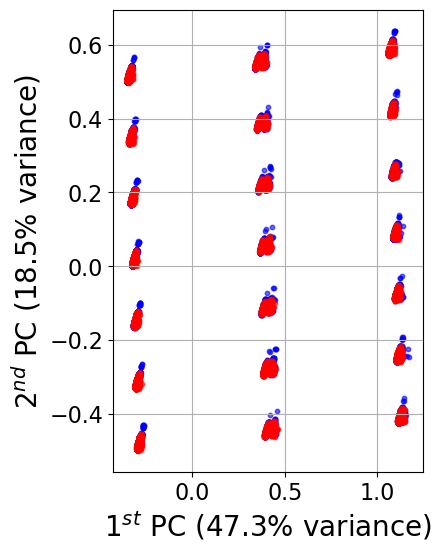} 
        \subcaption{\centering Experiment 3}
        \label{fig:pca_exp_3}
    \end{minipage}
    \caption{\centering \textit{TVAE} - PCA analysis of real (blue) vs. synthetic (red) flight information.}
    \label{fig:pca_exp_1_2_3}
\end{figure}

The performance of TVAE exhibited significant dependency on feature selection and data type representation across three experimental configurations: \textit{``df\_utc\_ts"}, \textit{``df\_utc\_d"} and \textit{``df\_utc\_d\_2"} in Experiments 1, 2, and 3, respectively. This limitation is likely attributed to poor latent space learning, a known issue in VAEs referred to as posterior collapse \cite{Zhao2020DiscretizedBI, Tong2023InVAErtNA, Hao2023MetaOptimizedJG, He2019LaggingIN, Kuzina2023DiscouragingPC}. When temporal features were input in datetime format, the synthetic data failed to accurately replicate the statistical structure of the original data, as illustrated in Fig.~\ref{fig:pca_exp_1}.

\begin{figure}[H]
    \centering
    \begin{minipage}{0.155\textwidth}
        \centering
        \begin{minipage}{\textwidth}
            \centering
            \includegraphics[width=\textwidth]{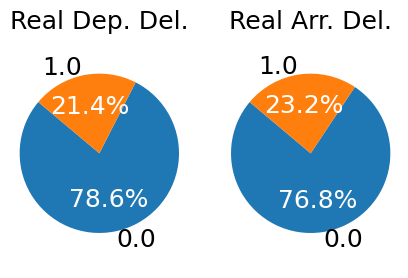} 
        \end{minipage}
        \begin{minipage}{\textwidth}
            \centering
            \includegraphics[width=\textwidth]{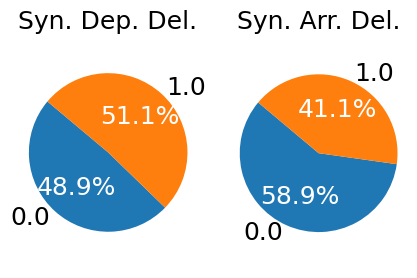} 
        \end{minipage}
    \subcaption{\centering Experiment 1}
    \label{fig:class_balance_exp_1}
    \end{minipage}
    \hfill
    \begin{minipage}{0.155\textwidth}
        \centering
        \begin{minipage}{\textwidth}
            \centering
            \includegraphics[width=\textwidth]{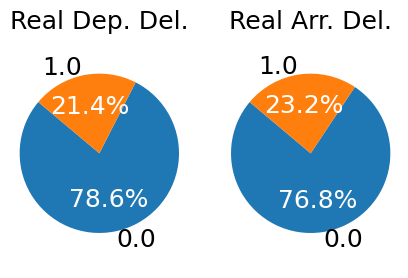} 
        \end{minipage}
        \begin{minipage}{\textwidth}
            \centering
            \includegraphics[width=\textwidth]{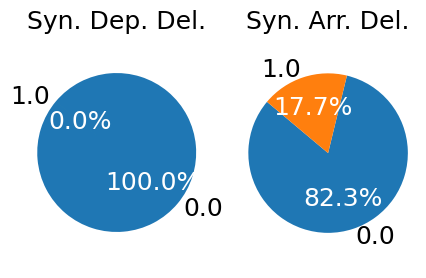} 
        \end{minipage}
    \subcaption{\centering Experiment 2}
    \label{fig:class_balance_exp_2}
    \end{minipage}
    \hfill
    \begin{minipage}{0.155\textwidth}
        \centering
        \begin{minipage}{\textwidth}
            \centering
            \includegraphics[width=\textwidth]{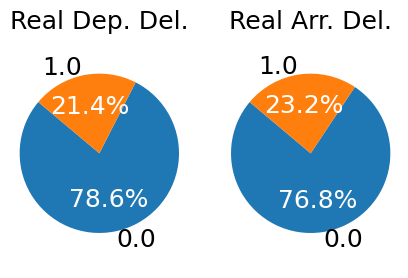} 
        \end{minipage}
        \begin{minipage}{\textwidth}
            \centering
            \includegraphics[width=\textwidth]{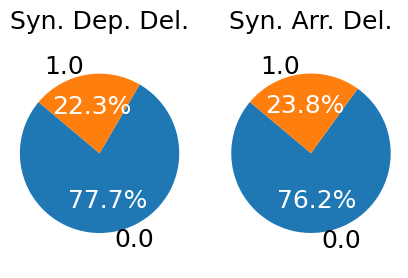} 
        \end{minipage}
    \subcaption{\centering Experiment 3}
    \label{fig:class_balance_exp_3}
    \end{minipage}
    \caption{\centering \textit{TVAE} - Class balance analysis of real (top) vs. synthetic (bottom) departure and arrival delay labels (1 = delayed, 0 = on time).}
    \label{fig:class_balance_exp_1_2_3}
\end{figure}

Replacing datetime features with numerical time duration values led to a partial improvement in capturing data variability. However, Fig.~\ref{fig:pca_exp_2} indicates that certain clusters present in the real dataset remain absent in the synthetic data. This observation is further supported by Fig.~\ref{fig:class_balance_exp_2}, which shows that the generated dataset contains no instances of delayed departures (0\%). These findings suggest that the generative model failed to learn and reproduce departure-related patterns, with the missing clusters in the PCA representation likely corresponding to departure-related information. To address this limitation, we refined the feature set by transitioning from \textit{``df\_utc\_d"} to \textit{``df\_utc\_d\_2"}, incorporating an additional temporal feature—``Actual Departure Time UTC"—at the departure airport. This modification aims to enhance the model's ability to capture departure-related patterns and improve the representation of delayed departures in the synthetic dataset. The impact of this adjustment is evident in Figs.~\ref{fig:pca_exp_3} and~\ref{fig:class_balance_exp_3}.

Another significant distinction between \textit{``df\_utc\_d"} and \textit{``df\_utc\_d\_2"} lies in the inclusion of the ``Taxi In Time (min)" feature. This feature can be calculated as the time difference between ``Actual Arrival Time UTC" and ``Wheels On Time UTC". However, while this feature depends on other time-related variables, it is also strongly influenced by the arrival airport's characteristics. Calculating it post-generation would only account for its temporal dependencies while largely neglecting its relationship with the arrival airport. Therefore, we explicitly incorporated it into \textit{``df\_utc\_d\_2"}, enabling the model to learn both the correlation between ``Taxi In Time (min)" and the arrival airport, as well as its relationships with other temporal features.

\begin{figure}[H]
    \centering
    \begin{minipage}{0.155\textwidth} 
        \centering
        \includegraphics[width=\textwidth]{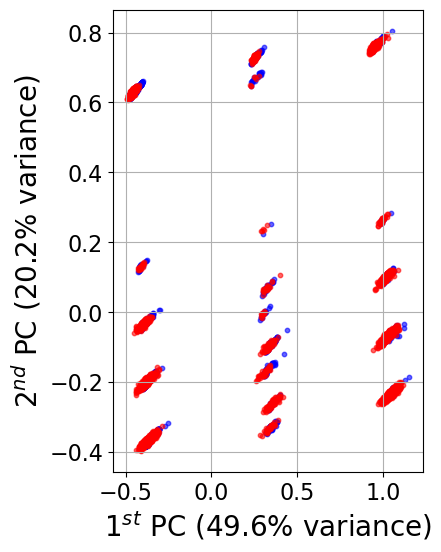} 
        \subcaption{\centering Experiment 4}
        \label{fig:pca_exp_4}
    \end{minipage}
    \begin{minipage}{0.155\textwidth} 
        \centering
        \includegraphics[width=\textwidth]{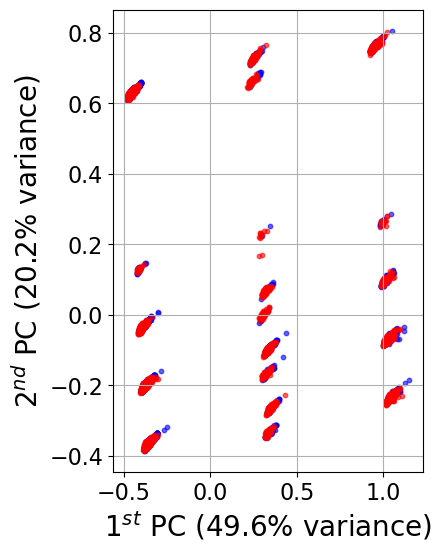} 
        \subcaption{\centering Experiment 5}
        \label{fig:pca_exp_5}
    \end{minipage}
    \caption{\centering \textit{GC} - PCA analysis of real (blue) vs. synthetic (red) flight information.}
    \label{fig:pca_exp_4_5}
\end{figure}

\begin{figure}[H]
    \centering
    \begin{minipage}{0.155\textwidth}
        \centering
        \begin{minipage}{\textwidth}
            \centering
            \includegraphics[width=\textwidth]{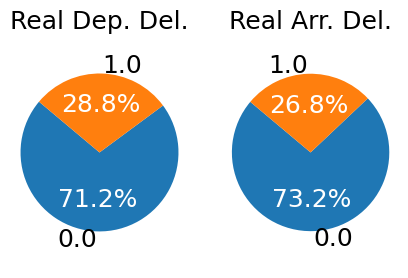} 
        \end{minipage}
        \begin{minipage}{\textwidth}
            \centering
            \includegraphics[width=\textwidth]{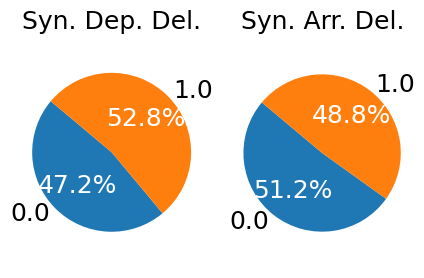} 
        \end{minipage}
    \subcaption{\centering Experiment 4}
    \label{fig:class_balance_exp_4}
    \end{minipage}
    \begin{minipage}{0.155\textwidth}
        \centering
        \begin{minipage}{\textwidth}
            \centering
            \includegraphics[width=\textwidth]{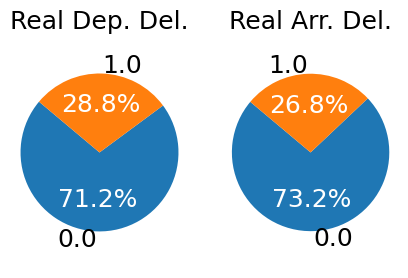} 
        \end{minipage}
        \begin{minipage}{\textwidth}
            \centering
            \includegraphics[width=\textwidth]{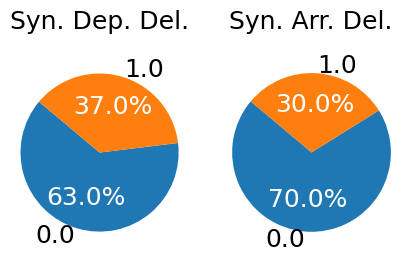} 
        \end{minipage}
    \subcaption{\centering Experiment 5}
    \label{fig:class_balance_exp_5}
    \end{minipage}
    \caption{\centering \textit{GC} - Class balance analysis of real (top) vs. synthetic (bottom) departure and arrival delay labels (1 = delayed, 0 = on time).}
    \label{fig:class_balance_exp_4_5}
\end{figure}

Unlike TVAE, the CG model demonstrated greater robustness to feature selection and data types across both experimental configurations: \textit{``df\_utc\_ts"} in Experiment 4 and \textit{``df\_utc\_d\_2"} in Experiment 5. As shown in Fig.~\ref{fig:pca_exp_4_5}, the synthetic flight data generated by CG effectively preserved the full variability of the real dataset, even when temporal features were represented in datetime format, as in Experiment 4. Furthermore, when using \textit{``df\_utc\_d\_2"} as input, the GC-generated synthetic data exhibited a class distribution more closely matching that of the real data, as illustrated in Fig.~\ref{fig:class_balance_exp_5}. The cluster patterns in Fig.~\ref{fig:pca_exp_4_5} differ from those in Fig.~\ref{fig:pca_exp_1_2_3}, due to the different data sizes used with TVAE and GC.

The diversity analysis demonstrated that using the \textit{``df\_utc\_d\_2"} DataFrame as input for both the TVAE and GC generative models resulted in improved diversity coverage and a class distribution more closely aligned with the real data. Consequently, the subsequent evaluation will focus exclusively on Experiments 3 and 5.

\subsection{Statistical Assessment}
\label{sec:statistical_assessment}
\begin{figure}[H]
    \centering
    \begin{minipage}{0.24\textwidth} 
        \centering
        \includegraphics[width=\textwidth]{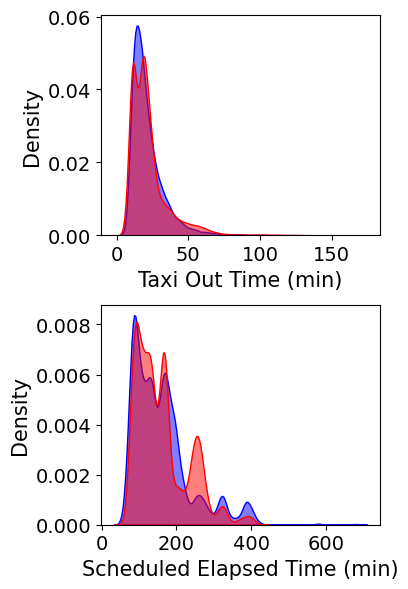} 
        \subcaption{\centering Experiment 3}
        \label{fig:stat_exp3}
    \end{minipage}
    \begin{minipage}{0.24\textwidth} 
        \centering
        \includegraphics[width=\textwidth]{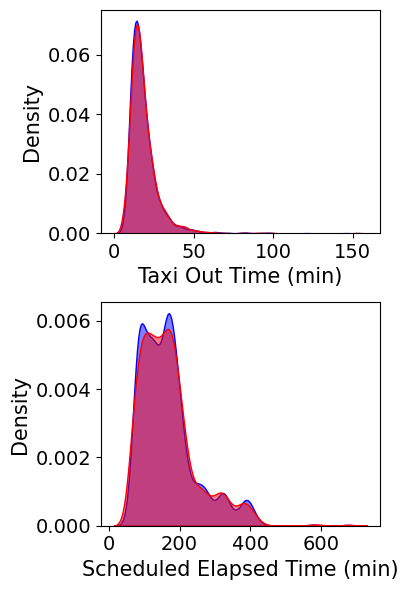} 
        \subcaption{\centering Experiment 5}
        \label{fig:stat_exp5}
    \end{minipage}
    \caption{\centering Similarity of marginal distributions for real (blue) vs. synthetic (red) data.}
    \label{fig:stat_exp_3_5}
\end{figure}

Both TVAE and GC generated synthetic flight data that closely matched the real data's individual feature distributions and pairwise feature relationships. Fig.~\ref{fig:stat_exp_3_5} illustrates this similarity by comparing the distributions of two features between real and synthetic data. The distributional differences observed between Fig.~\ref{fig:stat_exp3} and Fig.~\ref{fig:stat_exp5} stem from the varying training dataset sizes used for TVAE and GC, necessitated by computational constraints.

This statistical similarity was quantitatively validated through the Total Variation Distance (TVD) and Kolmogorov-Smirnov test results, which showed minimal divergence between the real and synthetic datasets across both categorical and numerical features. The Correlation Similarity and Contingency Similarity metrics confirmed that both generative models effectively preserved inter-feature statistical dependencies, as detailed in Table~\ref{tab:statistical}.

\renewcommand{\arraystretch}{1.2} 
\setlength{\tabcolsep}{8pt} 
\begin{table}[H]
\caption{\centering Quantitative evaluation of the statistical similarity between synthetic and real data.}
\begin{center}
\resizebox{0.47\textwidth}{!}{%
\begin{tabular}{||l||c|c||}
\hline \rowcolor{darkgray} \textbf{Statistical similarity} & \textbf{Experiment 3} & \textbf{Experiment 5} \\
\hline  \hline \cellcolor{darkgray} Marginal distribution & 87.53\% & 94.41\% \\
\hline \cellcolor{darkgray} Bivariate distributions & 75.41\% & 83.7\% \\
\hline \cellcolor{darkgray} Average & 81.47\% & 89.06\% \\
\hline
\end{tabular}
}
\label{tab:statistical}
\end{center}
\end{table}

The Gaussian Copula model demonstrated superior performance compared to TVAE, achieving higher accuracy in replicating the statistical properties of the real data. GC's enhanced capability was particularly evident in its reproduction of both univariate distributions and bivariate relationships, establishing it as the more effective model for replicating the statistical characteristics of the real flight data. 

\subsection{Fidelity Assessment}
\label{sec:fidelity_assessment}
As anticipated from the previous assessment, which demonstrated approximately 90\% similarity between the Gaussian Copula-generated synthetic data and real data, the seven selected classifiers showed reduced accuracy in distinguishing between real and GC-generated synthetic flight data in Experiment 5. Table~\ref{tab:fidelity} quantifies this performance, showing lower average accuracy (66.93\%) and F1 score (54.72\%) compared to Experiment 3, further validating the high quality of the GC-generated synthetic data.

\renewcommand{\arraystretch}{1.2} 
\setlength{\tabcolsep}{8pt} 
\begin{table}[H]
\caption{\centering Classification performance metrics for distinguishing real from synthetic flight data.}
\begin{center}
\resizebox{0.47\textwidth}{!}{%
\begin{tabular}{||l||c|c||}
\hline \rowcolor{darkgray} \textbf{Discriminative score} & \textbf{Experiment 3} & \textbf{Experiment 5} \\
\hline  \hline \cellcolor{darkgray} Average Accuracy & 78.39\% & 66.93\% \\
\hline \cellcolor{darkgray} Average F1 Score & 74.41\% & 54.72\% \\
\hline
\end{tabular}
}
\label{tab:fidelity}
\end{center}
\end{table}
\subsection{Utility Assessment}
\label{sec:utility_Assessment}
The regression models trained on TVAE-generated synthetic data in Experiment 3 achieved comparable or superior accuracy when tested on real data (TSTR), with mean absolute errors around 11 minutes for arrival delay predictions, as shown in Table~\ref{tab:utility}. However, despite producing statistically superior synthetic data that was harder to distinguish from real data, GC-generated data was less effective for training predictive models. This limitation arose from computational constraints that restricted GC's synthetic dataset to approximately 2,000 flights, compared to TVAE's 52,000 flights (Table~\ref{tab:data_size}).

To illustrate the impact of this difference, assume the real dataset contains 500 unique flight routes, with flights evenly distributed among them. Under this assumption, TVAE-generated data would provide 104 flights per route, whereas GC-generated data would yield only 4 flights per route. This small sample size limits the model’s ability to learn flight delay patterns and capture route-specific variability. While real-world distributions are more uneven, this simplified example highlights why the smaller GC dataset results in weaker predictive performance despite its stronger statistical resemblance to real data.

\renewcommand{\arraystretch}{1.2} 
\setlength{\tabcolsep}{8pt} 

\begin{table}[H]
\caption{\centering Predictive performance of machine learning models trained on real vs. synthetic flight data for arrival delay prediction.}
\begin{center}
\resizebox{0.47\textwidth}{!}{%
\begin{tabular}{||l||c|c|c|c||}

\hline \cellcolor{darkgray} & \multicolumn{2}{c|}{\cellcolor{darkgray}\textbf{Experiment 3}} & \multicolumn{2}{c||}{\cellcolor{darkgray}\textbf{Experiment 5}} \\ 

\arrayrulecolor{darkgray} 
\cline{1-5}
\arrayrulecolor{black} 
\hhline{|~|----|}

\multirow{-2}{*}{\cellcolor{darkgray}\textbf{Predictive score}} &  \cellcolor{darkgray}\textbf{TRTR} & \cellcolor{darkgray}\textbf{TSTR} & \cellcolor{darkgray}\textbf{TRTR} & \cellcolor{darkgray}\textbf{TSTR} \\

\hline  \hline \cellcolor{darkgray} Average RMSE & 15.50 & 14.72 & 12.89 & 20.11 \\
\hline \cellcolor{darkgray} Average MAE & 11.50 & 11.06 & 9.48 & 14.66 \\
\hline \cellcolor{darkgray} Average R² & 0.76 & 0.79 & 0.86 & 0.66 \\
\hline
\end{tabular}
}
\label{tab:utility}
\end{center}
\end{table}

\section{Discussion}
\label{sec:discussion}
\begin{figure}[H]
    \centering
    \includegraphics[width=0.48\textwidth]{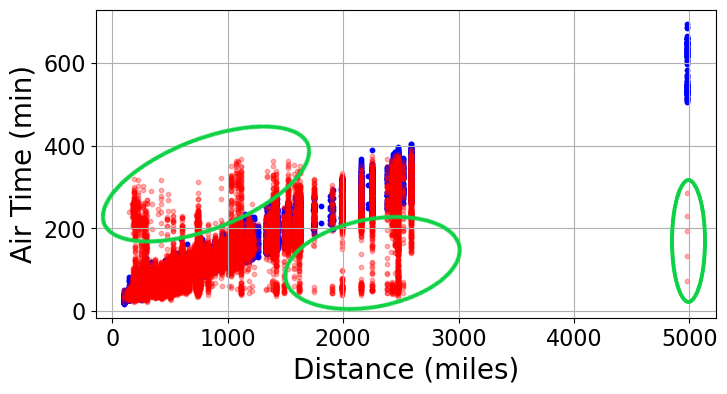} 
    \caption{\centering Correlation between distance and air time for real (blue) vs. synthetic (red) data.}
    \label{fig:corr}
\end{figure}

An important observation is that assessing only the marginal distributions or the statistical similarity of individual features is insufficient. It is equally essential to visually and numerically examine the joint distributions between pairs of variables. For instance, unlike air time, the distances between airport pairs are not explicitly provided as direct inputs to the generative model; instead, they are inferred post-generation based on airport ID pairs. As a result, the generator can only establish relationships between air time and airport ID pairs, rather than directly modeling the distance. By analyzing the correlation between distance and air time in Fig.~\ref{fig:corr}, it is evident that the synthetic data contains some incorrect air time values (highlighted in green circles) that are not proportional to the corresponding airport distances. To address this, the generation process should be refined to minimize the occurrence of these incorrect values, and any remaining inaccuracies should be filtered out during post-generation cleaning.

Some variation in the marginal distributions between real and synthetic data is acceptable and does not necessarily indicate an issue with the synthetic data. For instance, synthetic data might reflect a higher number of flights between airports $X_1$ and $Y_1$ (1000 miles apart) and fewer flights between airports $X_2$ and $Y_2$ (500 miles apart) compared to real data. This discrepancy would cause a shift in the peaks of the marginal distribution for the distance feature, showing fewer instances of 500-mile flights and more instances of 1000-mile flights. However, it is crucial that the bivariate distributions (i.e., the correlations between features) remain consistent between the real and synthetic data in order to maintain operational validity.

Another key takeaway is that although the Gaussian Copula (GC) model demonstrated higher statistical similarity and fidelity than the Tabular Variational Autoencoder (TVAE), it was less effective in terms of synthetic data utility. This underscores the importance of a multi-faceted evaluation framework. Furthermore, it highlights GC’s limitations in handling large datasets and the critical role that dataset size plays in capturing and accurately predicting flight delay patterns.  

\section{Conclusions \& Future Work}
\label{sec:conclusions_and_future_work}
In this study, we explored the use of generative models for producing realistic synthetic flight information and established a rigorous four-stage evaluation process to assess the statistical similarity, fidelity, diversity, and predictive utility of the synthetic data. While both TVAE and GC models demonstrated the ability to generate high-quality synthetic data, TVAE was sensitive to data types and feature selection, which affected its performance in certain cases. On the other hand, GC achieved higher statistical similarity and fidelity. However, GC’s computational limitations restricted its application to larger datasets, ultimately affecting the utility of the GC-generated data for predictive modeling. In contrast, TVAE was capable of handling larger datasets efficiently and, once trained, enabled fast and scalable sampling of synthetic data, making it more practical for large-scale applications.

Despite these limitations, our findings indicate that synthetic data can be effectively used to train flight delay prediction models, achieving accuracy comparable to models trained on real data. This brings us one step closer to providing the aviation community with an abundant source of reliable synthetic flight data, adaptable to different operational scenarios.

This study lays the foundation for future research on synthetic data generation methods specifically tailored to address the unique challenges of air transport applications. Future work will focus on refining the generative process by leveraging increased computational power to test GC on larger datasets, similar to TVAE. Additionally, strategies to mitigate posterior collapse in TVAE will be explored, along with an assessment of their impact on TVAE's sensitivity to data types and feature selection. Hyperparameter tuning will also be investigated to optimize feature correlations and improve the operational correctness of the generated data. Moreover, in this analysis, rejection sampling was applied to filter out synthetic routes that did not exist in historical data. Future research will analyze these newly generated routes to assess their plausibility and potential insights. Finally, the scope of this work will be expanded by incorporating additional flight attributes, such as diversions and cancellations, to further enhance the applicability of synthetic flight data in air transportation research.  


\section*{Acknowledgment}
This paper is based on the work done in the SynthAIr project. SynthAIr has received funding from the SESAR Joint Undertaking under the European Union’s Horizon Europe research and innovation programme under grant agreement No 101114847.  Views and opinions expressed are however those of the authors only and do not necessarily reflect those of the European Union or SESAR 3 Joint Undertaking. Neither the European Union nor SESAR 3 Joint Undertaking can be held responsible for them.

\bibliographystyle{IEEEtran}
\bibliography{References}
\end{document}